\documentclass[conference]{IEEEtran}
\IEEEoverridecommandlockouts
\usepackage{cite}
\usepackage{amsmath,amssymb,amsfonts}
\usepackage{graphicx}
\usepackage{textcomp}
\usepackage{xcolor}

\usepackage{times}  % DO NOT CHANGE THIS
\usepackage{helvet}  % DO NOT CHANGE THIS
\usepackage{courier}  % DO NOT CHANGE THIS
\usepackage[hyphens]{url}  % DO NOT CHANGE THIS
\usepackage{graphicx} % DO NOT CHANGE THIS

\usepackage{latexsym}

\usepackage{booktabs}
\usepackage{mathtools}
\usepackage{varwidth}
\usepackage{empheq}
\usepackage{amssymb}
\usepackage{tabularx}
\usepackage{multirow}
\usepackage{epsfig}
\usepackage{amsmath}
\usepackage{cancel}
\usepackage{amsfonts}       % blackboard math symbols
\usepackage{nicefrac}       % compact symbols for 1/2, etc.
\usepackage{bbm}
\usepackage{bm}
\usepackage{defs}
\usepackage{enumitem}% http://ctan.org/pkg/enumitem

\renewcommand{\vec}[1]{\boldsymbol{#1}}
\renewcommand{\vec}[1]{\boldsymbol{#1}}

\usepackage{siunitx} % to round numbers
\renewcommand{\vec}[1]{\boldsymbol{#1}}
\usepackage{algorithm}
\usepackage{algpseudocode}

\usepackage{microtype}

\usepackage{caption} % DO NOT CHANGE THIS AND DO NOT ADD ANY OPTIONS TO IT
\usepackage{subcaption}
\DeclareCaptionStyle{ruled}{labelfont=normalfont,labelsep=colon,strut=off} % DO NOT CHANGE THIS
\usepackage{algorithm}

\usepackage{newfloat}
\usepackage{listings}
\floatstyle{ruled}
\newfloat{listing}{tb}{lst}{}
\floatname{listing}{Listing}

\def\BibTeX{{\rm B\kern-.05em{\sc i\kern-.025em b}\kern-.08em
    T\kern-.1667em\lower.7ex\hbox{E}\kern-.125emX}}
\begin{document}

\title{Gradient Sparsification For \emph{Masked Fine-Tuning} of Transformers}

\author{\IEEEauthorblockN{1\textsuperscript{st} James O' Neill}
\IEEEauthorblockA{\textit{Huawei Ireland Research Center} \\
Dublin, Ireland \\
james.o.neil@huawei-partners.com}
\and
\IEEEauthorblockN{2\textsuperscript{nd} Sourav Dutta}
\IEEEauthorblockA{\textit{Huawei Ireland Research Center} \\
Dublin, Ireland \\
sourav.dutta2@huawei.com}
}

\maketitle

\begin{abstract}
Fine-tuning pretrained self-supervised language models is widely adopted for transfer learning to downstream tasks. Fine-tuning can be achieved by freezing gradients of the pretrained network and only updating gradients of a newly added classification layer, or by performing gradient updates on all parameters. Gradual unfreezing makes a trade-off between the two by gradually unfreezing gradients of whole layers during training. This has been an effective strategy to trade-off between storage and training speed with generalization performance. However, it is not clear whether gradually unfreezing layers throughout training is optimal, compared to sparse variants of gradual unfreezing which may improve fine-tuning performance. In this paper, we propose to stochastically mask gradients to regularize pretrained language models for improving overall fine-tuned performance. We introduce GradDrop and variants thereof, a class of gradient sparsification methods that mask gradients during the backward pass, acting as gradient noise. GradDrop is sparse and stochastic unlike gradual freezing. Extensive experiments on the multilingual XGLUE benchmark with XLMR-Large show that GradDrop is competitive against methods that use additional translated data for intermediate pretraining and outperforms standard fine-tuning and gradual unfreezing. A post-analysis shows how GradDrop improves performance with languages it was not trained on, such as under-resourced languages.

\iffalse
Cross-lingual language modeling has shown to be an effective pretraining strategy for learning representations that generalize well to downstream tasks they are used for. However, fine-tuning to a set of tasks require fine-tuning of multiple linear layers that use the last hidden representation of the pretrained model as input. In this work, we improve the efficiency of fine-tuning for pretrained cross-lingual models by learning which layers gradients to freeze and unfreeze. 
% to mask the weights that are irrelevant for the specific set of tasks while fine-tuning the linear layer towards them. 
Experiments on GLUE, XNLI and unsupervised machine translation show that our \emph{supermasking} strategy consistently outperforms baselines. 
\fi

\end{abstract}


%%%%%%%%% BODY TEXT
\section{Introduction}
Fine-tuning pretrained transformer models for downstream tasks has been the defacto standard in natural language processing due to the recent successes of large-scale masked language modeling~\cite{radford2018improving,Devlin2019bert,lample2019cross,Conneau2020xlm-r}. This is usually achieved in one of two ways: (1) freeze the gradients of the pretrained portion of the network and perform stochastic gradient descent (SGD) on a newly added task-specific layer/s or (2) perform SGD on both the pretrained and newly added layer/s. However, freezing all gradients of the pretrained layers can be too restrictive, particularly when the downstream task is dissilmilar to the task of language modeling used during pretraining~\cite{peters2019tune}. In contrast, unfreezing all layers may lead to negative transfer whereby irrelevant features are tuned for a downstream task or stability issues may arise when performing stochastic gradient descent for a large number of parameters~\cite{liu2020understanding}. 
\newline
Gradual unfreezing~\cite{howard2018universal} is an alternative method that only tunes a subset of $k$ layers and freezes the remaining layers for an epoch. In each successive epoch, the next subset of $k$ layers are tuned and this is iterated in a top-down fashion, i.e., unfreeze top $k$ subset of layers to the bottom $k$ subset of layers. Gradual unfreezing reduces training time by reducing the number of gradient updates after backpropagation. For the sole purpose of improving fine-tuning performance, gradual unfreezing could benefit from sparse gradient dropout alternatives that allow at least a subset of weights of all layers to be tuned at each epoch. Concretely, instead of restricting the freezing of gradients for \emph{whole} layers, we can mask a percentage of gradients in \emph{all} layers to allow gradients to flow through the whole network.

Thus, in this paper we propose \emph{gradient dropout}, which we refer to as {\em GradDrop}, for {\em stochastically masking gradients} to regularize pretrained language fine-tuning. We find two particular variants of GradDrop significantly improve the fine-tuning of pretrained models, namely {\em GradDrop-Epoch} (where weight masks are fixed over the whole epoch) and {\em Layer-GradDrop} (where we stochastically masks out gradients of whole layers). Our experiments focus cross-lingual language model (LM) XLM-R$_{\text{Large}}$~\cite{conneau2019unsupervised}, given its wide adoption and success in transfer learning to various languages.
%as follows.
\vspace{-0.5em}
\subsection{Contributions} 
In summary, our main contributions are: 
\begin{enumerate}[topsep=0pt,itemsep=-1ex,partopsep=1ex,parsep=1ex]
\itemsep0em 
\item A dropout variant called \emph{gradient dropout} (\emph{GradDrop}) that regularizes fine-tuned models by randomly removing gradients during training. We also propose a variant,  (\emph{GradDrop-Epoch}) that updates the gradient mask every epoch instead of every mini-batch. GradDrop and its variants are \emph{simple to implement} and thus can be used with little effort as a default operation for LM fine-tuning. 
\item Stochastic gradual unfreezing whereby layers are chosen at random for gradient updates at each epoch. We refer to this as {\em Layer-GradDrop} and compare this to standard fine-tuning (SFT) and gradual unfreezing. 
\item A comprehensive analysis of the how masking and fine-tuning can be used to improve cross-lingual transfer to downstream tasks without any \textit{task-specific cross-lingual alignment} or \textit{translate-train} training schemes. 
\end{enumerate}

\section{Related Research}
Before discussing our proposed regularizer, we review existing approaches to LM fine-tuning, cross-lingual LM fine-tuning and other methods that have explored masking strategies on pretrained LMs. 

\subsection{Language Model Fine-Tuning}

% While the predominant methodology for transfer learning is to fine-tune all weights of the pre-trained model, adapters have recently been introduced as an alternative approach in the NLP domain ~\cite{Houlsby2019adapters,Bapna2019adapters,pfeiffer2020AdapterHub,pfeiffer2020adapterfusion,Wang:2020iclr}. 

\textbf{Adapters} have shown success by fine-tuning relatively small linear layers, referred to as \emph{bottlenecks}, that are placed between pretrained frozen layers and generally only account for a small percentage (e.g., 2-5\%) of the overall number of parameters in the pretrained model. There are variants whereby some adapters are placed only on the output of each self-attention block, within each self-attention block, or combining adapters that have been independently trained for specific tasks and languages~\cite{pfeiffer20madx}. Current work predominantly focuses on training adapters for each task separately~\cite{Houlsby2019adapters,pfeiffer2020adapterfusion,pfeiffer2020AdapterHub}, which enables parallel training and subsequent combination of the weights. 
% In NLP, adapters have been mainly used within deep transformer-based architectures~\cite{vaswani2017attention}. 
% \textbf{Structured Dropout}~\citet{fan2019reducing} drop structured modules (e.g self-attention blocks) during training such that a portion of self-attention heads are randomly dropped out during training. This leads to improved regularization and is robust to pruning whole blocks after training. We consider \emph{LayerDrop} as a baseline, particularly for comparison with our proposed GradDrop.  
% \textbf{AdapterDrop}
Ruckle et al.~\cite{ruckle2020adapterdrop} 
remove adapters from lower layers during training and inference, incorporating structured dropout~\cite{fan2019reducing} with adapters~\cite{Houlsby2019adapters}. This leads to parameter reduction while maintaining task performances, with further improvements when pruning adapters using Adapter Fusion~\cite{pfeiffer2020adapterfusion}.

\paragraph{Gradual Unfreezing}
%  by freezing or unfreezing the gradients of the pretrained portion of the network and fine-tune a new linear layer on top for a downstream task of interest. However, the gap between the language model training task (e.g masked language modeling) and the chosen downstream task can lead to instability in fine-tuning or negative transfer learning whereby a significant subset of the parameters are irrelevaent for the downstream task. Gradual (gradient) unfreezing~\cite{howard2018universal} addresses the aforementioned problems by reducing the number of layers fine-tuned at each training epoch to a small subset of the total number of layers.
Howard et al.~\cite{howard2018universal} proposed gradually turning on gradients layer by layer for LM pretraining and fine-tuning, leading to a reduction in training time due to a reduction in gradient updates. Peters et al.~\cite{peters2019tune} have further explored which tasks benefit from fine-tuning when all gradients are active, when only the newly added fine-tuning layer gradients are active and when using gradual unfreezing. Their main finding is that when the underlying LM pretraining is semantically similar to the downstream task there is less need to deactivate gradients, while the semantically different tasks benefit more from activating all gradients for fine-tuning. 

\subsubsection{Cross-Lingual Fine-Tuning}
Ren et al.~\cite{ren2019explicit} use cross-lingual pretraining to improve performance on unsupervised neural machine translation (UNMT) by computing cross-lingual n-gram embedding and predicting an n-gram translation table from them. From this, they introduce cross-lingual MLM where they sample n-grams for a given input text and predict the translation n-grams at each time step. 

Muller et al.~\cite{muller2021first} show that multilingual BERT (mBERT), a popular multilingual LM, can be viewed as the stacking of a multilingual encoder followed by a task-specific language-agnostic predictor. While the encoder is crucial for cross-lingual transfer and remains mostly unchanged during fine-tuning, the task predictor has little importance on the transfer and can be reinitialized during fine-tuning.

% propose a fine-tuning method that improves fine-tuning of multilingual models for a specific language downstream task, referred to as multi-lingual LM fine-tuning (MultiFiT). 
Eisenschlos et al.~\cite{eisenschlos2019multifit} perform multi-lingual LM fine-tuning (MultiFiT) by combining universal language
model fine-tuning~\cite{howard2018universal} with quasi-recurrent neural network~\cite{bradbury2016quasi}, subword tokenization~\cite{kudo2018sentencepiece} and a cross-lingual LM teacher network to distill the monolingual fine-tuned model to the zero-shot setting. 
Fang et al.~\cite{fang2020filter} improve cross-lingual fine-tuning by first performing cross-lingual alignment, prior to downstream fine-tuning, by first learning language independent representations which are then concatenated and passed as input to another self-attention block that learns the cross-lingual features. 
%They show that learning language independent features prior to learning cross-lingual features performs better than only using one or the other. 

% For simple tasks such as classification, translated text in the target language shares the same label as the source language. However, this shared label becomes less accurate or even unavailable for more complex tasks such as question answering, NER and POS tagging. To tackle this issue, we further propose an additional KL-divergence self-teaching loss for model training, based on auto-generated soft pseudo-labels for translated text in the target language. Extensive experiments demonstrate that FILTER achieves new state of the art on two challenging multilingual multi-task benchmarks, XTREME and XGLUE

\subsection{Pretrained Model Masking}
While standard LM fine-tuning remains the defacto standard in NLP-based transfer learning, there has been other masking-related approaches. Zhao et al.~\cite{zhao2020masking} have learned a mask over the weights instead of fine-tuning the weights, showing that this can lead to competitive performance for fine-tuning. In contrast to our work, we show that combining masking during fine-tuning is a preferred method for the same computational budget.
Liu et al.~\cite{liu2021autofreeze} use the change of the gradient magnitudes of a layer as a criterion to determine whether a layer is to be frozen. Hence, gradients that stagnate in a layer are most likely to be frozen during the fine-tuning process.%, leading to a reduction in the number of gradient updates and in turn, faster training time. 
Chen et al.~\cite{chen2020just} have explored the problem of conflicting gradient signs in the multi-task setting where multiple gradients are assigned to a single weight for each task. To avoid conflicting gradient signs, they choose a single gradient and mask the remaining gradients based on the gradient distribution for each weight. In contrast to our GradDrop, we do not focus on this multi-task setting and our masking refers to 1) zeroing gradients out for a weight (i.e binary) and not choosing a gradient among many (real-valued) and 2) we only have a single gradient for single task training as opposed to the gradient distribution in the multi-task setting. 

, not according to a distribution of multi-task gradient signals

% In contrast to our work, we aim to reduce by both training and inference time using gradient sparsification techniques.  

\section{Proposed Methodology}

In this section, we describe our main contribution, gradient dropout and variants thereof. We begin by first describing the self-attention blocks in transformers. 
Assume we have a sequence of vectors $\vec{x}_1, \ldots, \vec{x}_n$ where each vector $\vec{x}_i \in \mathbb{R}^{d}$ of $d$ dimensions (e.g., $d=512$). We define $\mat{Q} \in \mathbb{R}^{n \times d}$ to be a matrix representing the sequence where the i-th row of $\mat{Q}$ corresponds to $\vec{x}_i$. The key $\mat{K} \in \mathbb{R}^{d \times l}$, value $\mat{V} \in \mathbb{R}^{d \times l}$ and projection layer $\mat{U} \in \mathbb{R}^{d \times o}$ parameters are defined where $\mat{U}$ ensures the output dimensionality of the self-attention block is the same as the original input $\mat{Q}$. We can then define the self-attention as Equation \eqref{eq:self_att_1},
\begin{equation}\label{eq:self_att_1}
\mat{Z} = \text{Softmax}\Bigg(\frac{\mat{Q}\mat{K}}{\sqrt{dl}}\mat{V}^{\top}\mat{Q}^{\top}\Bigg)\mat{Q}\mat{U}    
\end{equation}
where $\mat{Q}\mat{U} \in \mathbb{R}^{n \times o}$ is matrix of new embeddings, $\mat{Q}\mat{K}\mat{V}^{\top}\mat{Q}^{\top} \in \mathbb{R}^{n \times n}$ is a matrix representing the inner products in a new $l$-dimensional space and $\mathrm{Softmax}\big(\mat{Q}\mat{K}\mat{V}^{\top}\mat{Q}^{\top}\big)$ is a matrix where each row entry is positive and sums to 1. Note that \emph{scaled} dot-product is used (normalization by $\sqrt{dl}$) to avoid vanishing gradients of the $\mathrm{Softmax}$, which may occur when $dl$ is large. 

The parameters for the $j$-th attention head $\mat{K}^j, \mat{V}^j \in \mathbb{R}^{d \times l}$, $\mat{U}^j \in \mathbb{R}^o$ for $j = 1,\ldots, n_a$ where $n_a$ is the number of attention heads. Then we summarize the formulation of multi-headed self-attention as Equation \eqref{eq:multi_head_att},
\begin{equation}\label{eq:multi_head_att}
\begin{split}
\mat{Z}^j & = \text{Softmax}\Big(\frac{\mat{Q}\mat{K}^j}{\sqrt{dl}}(\mat{V}^j)^{\top}\mat{Q}^{T}\Big)\mat{Q}\mat{U}^j \\
\tilde{\mat{Z}} & = \mathrm{Concat}(\mat{Z}^1, \ldots \mat{Z}^n_a) \\
\mat{Z} & = \mathrm{Feedforward}(\mathrm{LayerNorm}(\tilde{\mat{Z}} + \mat{Q}))    
\end{split}
\end{equation}
where $\mat{Z}^{j} \in \mathbb{R}^{n \times d_{a}}$ and $\tilde{\mat{Z}} \in \mathbb{R}^{n \times d_{a}n_{a}}$, with $d_{a}$ being the dimensionality of the self-attention output. The above formulation omits positional embeddings which are learned embeddings that output representations that reflect sequential information in its inputs (i.e., the same token with different contexts won't have the same representations) by computing distance between token positions. Given this background, we now describe gradient dropout.

\vspace{-0.5em}
\subsection{Gradient Dropout}
After backpropogation, we apply a random binary mask on the gradients of $\mat{K}, \mat{V}$ and $\mat{U}$. For simplicity, let us assume $\mathbf{\theta}:= \{\mat{K}, \mat{V}, \mat{U}\}$ and the gradients of $\mathbf{\theta}$ are represented as $\mat{g}:=\nabla_{\mathbf{\theta}} \cL^s (f_{\theta}(\mat{Q}), \mat{Y})$, where $\mat{Y} \in \mathbb{N}^{n\times d}$ represents one-hot targets of dimension $d$. A binary mask $\mat{m}$ is then generated from a predefined distribution (e.g., Bernoulli or Gaussian) and applied over the gradients. The gradient update rule with gradient dropout can then be expressed as,
\begin{equation}\label{eq:mask_grad}
\mathbf{\theta}'_{l} = \mathbf{\theta}_{l} - \alpha * \mat{g}_{l} \odot \mat{m}_{l}   
\end{equation}
where $\alpha$ is the learning rate, $\odot$ performs the Hadamard product (i.e., the element-wise product of tensors) and $l \in L$ is the layer index. Given that the stochastic noise induced by SGD through random mini-batch training regularizes DNNs, we too expect that the random dropping of gradients will have a similar regularization effect. When $\mat{m}$ is generated from a Bernoulli distribution, we randomly zero the gradient with probability $p$, in which the process of sampling $m$ is formulated as:
\begin{equation}
\mat{m} \sim \frac{\mathrm{Bernoulli}(1 - p)}{1 - p}
\end{equation}
where the denominator $1 - p$ is the normalization factor. Note that, different from Dropout~\cite{srivastava2014dropout} which randomly drops the intermediate activations in a supervised learning network under a single task setting, we perform the {\em dropout on the gradient level}. We focus on binary masks for $\mat{m}$ as it is computationally efficient to generate and store low precision boolean tensors, in comparison to continuous noise such as the Gaussian distribution. Lastly, when applying gradient dropout layerwise (\textbf{Layer-GradDrop}), $\vec{m} \in \{0, 1\}^{L}$ where $l$-th element in $\mat{m}$ corresponds to whether that layers gradients are activated or not. When $\vec{m}_l=1$, a one matrix $\mathbf{1}$ of the same dimensionality as $\mat{g}_l$ is applied, and and zeros when $\vec{m}_l=0$\footnote{Please see the supplementary material for a pseudocode example of GradDrop used with XLM-R.}.
We posit that the main generalization benefits given by sparsely freezing gradients can be explained by how it slows down the total amount of gradient flow for each consecutive mini-batch during fine-tuning. This is particularly important for tasks that are more distant from the original self-supervised pretraining objective used prior to fine-tuning, i.e., converging too fast on a distant task may lose the generalization benefits given by the pretrained state.

\subsection{Epoch-wise Gradient Dropout}
We also propose a variant of GradDrop whereby the same dropout mask is applied to all mini-batches for a single epoch. The mask can be reset for successive epochs by uniformly sampling from the aforementioned Bernoulli distribution at the same dropout rate as before.
\begin{figure}[ht]
\centering
\includegraphics[scale=0.485]{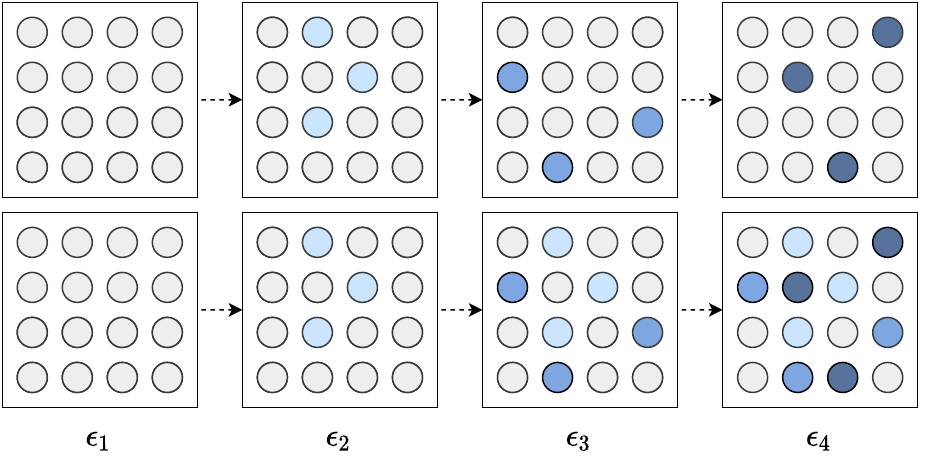}
\caption{\textbf{GradDrop-Epoch-Toggle (Top) and GradDrop-Epoch (Bottom)}. Grey represents frozen gradients, blue represents active gradients where darker blue indicates the recency of gradients turned on.}
\label{fig:epoch_gradmask}
\end{figure}
However, we also consider an accumulative mask whereby we sample from the Bernoulli distribution \emph{without} replacement for each epoch and this is the version we use for our experiments. 
Figure~\ref{fig:epoch_gradmask} shows the difference between the proposed {\em GradDrop-Epoch} when previous epoch masks are frozen once a new mask is applied (\textbf{GradDrop-Epoch-Toggle}) and when the previous epoch masks are left unfrozen (\textbf{GradDrop-Epoch}). In both cases, sampling without replacement is used, unlike standard GradDrop and like gradual unfreezing. This similarity to gradual unfreezing w.r.t. sampling without replacement aims to improve the stability during fine-tuning as only a subset of parameters are being updated for a whole epoch. Transformers are known to be difficult to train due to 
instability in optimization from their dependency on the residual branches within the self-attention blocks, as it amplifies parameter updates leading to larger changes to the model output~\cite{liu2020understanding}. These amplifications can be mitigated by stochastically freezing large portions of the network during fine-tuning when using GradDrop-Epoch, while allowing some gradient flow throughout all layers. 
% While layer normalization has mitigated this problem to some degree,

\subsection{Annealed Variants of Gradient Dropout} Thus far we have assumed all gradient dropout variants (GDVs) to have a fixed uniform gradient dropout rate throughout training. We can also apply each mask per minibatch or per epoch using a scheduled dropout rate that is non-uniform, such as exponential decay or a linear decay. In this work, we focus on a linear schedule that begins at $p=0.9$, reduces by $p_{\epsilon}:= p_{\epsilon-1} - 1/T$ at each epoch $\epsilon$ until the last epoch $T$ is reached where $p=0$. In subsequent tables, models that have term ``Anneal-'' use this annealed GradDrop schedule. 

\begin{table*}[ht]
\centering
\resizebox{\linewidth}{!}{
\begin{tabular}{l|ccccccccccccccc|c}
\toprule
Model & \textbf{en} & \textbf{ar} & \textbf{bg} & \textbf{de} & \textbf{el} & \textbf{es} & \textbf{fr} & \textbf{hi} & \textbf{ru} & \textbf{sw} & \textbf{th} & \textbf{tr} & \textbf{ur} & \textbf{vi} & \textbf{zh} & \textbf{Avg.} \\

\midrule
\textbf{Original XLM-R} & & & & & & & & & & & & & & & & \\
\midrule
XLM-R$_{\text{Base}}$~\cite{Conneau2020xlm-r} & 84.6 & 78.4 & 78.9 & 76.8 & 75.9 & 77.3 & 75.4 & 73.2 & 71.5 & 75.4 & 72.5 & 74.9 & 71.1 & 65.2 & 66.5 & 74.5 \\
XLM-R$_\text{Large}$~\cite{Conneau2020xlm-r} & 88.8 & 83.6 & 84.2 & 82.7 & 82.3 & 83.1 & 80.1 & 79.0 & 78.8 & 79.7 & 78.6 & 80.2 & 75.8 & 72.0 & 71.7 & 80.1 \\

\midrule
\textbf{FILTER}\iffalse~\cite{fang2020filter}\fi & & & & & & & & & & & & & & & & \\
%mBERT  & 80.8 & 64.3 & 68.0 & 70.0 & 65.3 & 73.5 & 73.4 & 58.9 & 67.8 & 49.7 & 54.1 & 60.9 & 57.2 & 69.3 & 67.8 & 65.4 \\
%MMTE                      & 79.6 & 64.9 & 70.4 & 68.2 & 67.3 & 71.6 & 69.5 & 63.5 & 66.2 & 61.9 & 66.2 & 63.6 & 60.0 & 69.7 & 69.2 & 67.5 \\
\midrule
%XLM$_{\text{Base}}$~\cite{fang2020filter} & 82.8 & 66.0 & 71.9 & 72.7 & 70.4 & 75.5 & 74.3 & 62.5 & 69.9 & 58.1 & 65.5 & 66.4 & 59.8 & 70.7 & 70.2 & 69.1 \\
XLM-R$_{\text{Large}}$~\cite{fang2020filter} & 88.7 & 77.2 & 83.0 & 82.5 & 80.8 & 83.7 & 82.2 & 75.6 & 79.1 & 71.2 & 77.4 & 78.0 & 71.7 & 79.3 & 78.2 & {79.2} \\
\shortstack{XLM-R$_{\text{Large}}$ (\emph{translate-train})}  & 88.6 & 82.2 & 85.2 & 84.5 & 84.5 & 85.7 & 84.2 & 80.8 & 81.8 & 77.0 & 80.2 & 82.1 & 77.7 & 82.6 & 82.7 & 82.6 \\

\midrule
\textsc{Filter} & 89.7 & 83.2 & 86.2 & 85.5 & 85.1 & 86.6 & 85.6 & 80.9 & 83.4 & 78.2 & 82.2 & 83.1 & 77.4 & 83.7 & 83.7 & 83.6 \\
\textsc{Filter} + Self-Teaching & 89.5 & 83.6 & 86.4 & 85.6 & 85.4 & 86.6 & 85.7 & 81.1 & 83.7 & 78.7 & 81.7 & 83.2 & 79.1 & 83.9 & 83.8 & 83.9 \\

\midrule
\textbf{Ours} & & & & & & & & & & &  & & & & & \\
\midrule

\shortstack{XLM-R}$_{\text{Large}}$ & 88.35 & 76.51 & 82.01 & 83.13 & 80.12 & 84.54 & 82.61 & 75.22 & 78.07 & 71.00 & 77.35 & 78.63 & 71.85 & 79.72 & 79.64 & {79.25} \\
\midrule
+GradFreeze-TopBottom & 88.83 & 77.83 & 80.76 & 83.25 & 80.73 & 84.46 & 83.22 & 74.18 & 79.24 & 72.05${\dagger}$ & 76.10 & 77.59 & 70.52 & 80.03 & 79.44 & 79.23 \\

+GradFreeze-BottomUp & 84.95 & 75.15 & 78.49 & 82.10 & 80.03 & 83.88 & 81.02 & 74.58 & 78.36 & 72.05${\dagger}$ & 75.83 & 77.08 & 69.17 & 79.43 & 79.01 & 78.07 \\

%\textsc{XLM-R}+Annealed GF-BT & \textbf{1} & 2 & 3 & 4 & 5 & \textbf{6} & 7 & 8 & 9 & 10 & \textbf{11} & 12 & 13 & 14 & 15 & 16 \\
%\textsc{XLM-R}+Annealed GF-TB & \textbf{1} & 2 & 3 & 4 & 5 & \textbf{6} & 7 & 8 & 9 & 10 & \textbf{11} & 12 & 13 & 14 & 15 & 16 \\
\midrule

% +GradDrop & 90.01 & 78,71 & 87.35 & 85.41 & 88.23 & 85.22 & 83.82 & 77.01 & 80.83 & 74.26 & 78.00 & 81.37 & 73.21 & 79.52 & 79.43 & 79.61 \\

+GradDrop & 90.01${\dagger}$ & 78.19 & 82.37${\dagger}$ & 83.53 & 80.68 & 84.82 & \textbf{83.69} & 76.18 & 78.72 & \textbf{72.73} & 77.03 & 79.04${\dagger}$ & \textbf{72.69} & \textbf{80.68} & 79.23 & {\textbf{79.97}} \\

+Anneal-GradDrop & 88.49 & 76.88 & 82.81 & 83.54${\dagger}$ & 80.13 & 85.07 & 82.90 & \textbf{78.11} & 78.14 & 71.04 & 76.72 & 78.39 & 72.47 & 80.17 & 79.28 & 79.58 \\

+Anneal-Layer-GradDrop & \textbf{90.68} & 78.19 & \textbf{82.93} & \textbf{83.57} & 80.96 & 85.26${\dagger}$ & 83.53${\dagger}$ & 76.27 & 79.12${\dagger}$ & 71.93 & 77.43${\dagger}$ & 77.59 & 72.49${\dagger}$ & 79.44 & \textbf{79.72} &  79.94$^{\dagger}$\\

% +Anneal-Layer-GradDrop & 88.73 & 77.04 & 82.48 & 83.09 & 80.82 & 84.99 & 82.34 & 77.51 & 78.47 & 70.19 & 77.25 & 78.84 & 72.07 & 80.01 & 79.97 & 79.58 \\

+Layer-GradDrop & 88.65 & 76.97 & 81.99 & 81.43 & 81.38 & 83.11 & 82.99 & 76.45 & \textbf{80.30} & 68.53 & \textbf{78.23} & 78.05 & 71.47 & 79.38 & 79.42${\dagger}$ & 79.22 \\

+GradDrop-Epoch & 88.27 &  \textbf{82.77} &  83.13 &  81.25 &  \textbf{88.71} & \textbf{85.30} &  83.25 &  77.07${\dagger}$ &  78.67 &  71.45 &  77.31 &  \textbf{79.40} &  72.53 &  80.20${\dagger}$ &  \textbf{79.72} &  79.94$^{\dagger}$ \\

\bottomrule
\multicolumn{17}{l}{The best performance obtained are marked in {\bf bold}, while the second best results are indicated with ${\dagger}$.}
\end{tabular}
}
\caption{XNLI zero-shot accuracy (apart from `en') for each language. Results of fine-tuned XLM-R from prior work~\cite{Conneau2020xlm-r,fang2020filter} are from the XTREME benchmark~\cite{hu2020xtreme}.}
\label{tbl:xnli_detailed_results}
%\vspace{-1mm}
\end{table*}

\section{Experimental Details}
In our experiments, we focus on cross-lingual tasks from the XGLUE benchmark~\cite{liang2020xglue}. For all tasks, we only use English language training data for fine-tuning XLM-R$_{\text{Large}}$ and evaluate ``zero-shot'' test performance on multiple other languages (not seen during training for fine-tuning) for each of the respective task.
We do not use \emph{any} cross-lingual alignment as a pretraining step and we also do not carry out the \textit{translate-train} fine-tuning scheme, 
%(denoted with ``-T'' in subsequent tables), 
which first translates all languages to a well-resourced target language such as English and then fine-tunes on the downstream task. This is because our aim is to be competitive against both cross-lingual alignment and translate-train based fine-tuning, as in many cases aligned or unaligned text is not easily available. 

For all GDVs, we apply gradient dropout to every layer apart from the input embedding layers and the task-specific classification layer that is on top of XLM-R$_{\text{Large}}$. We ensure that there is no dropout used on the layers when using gradient dropout and when not using gradient dropout in SFT, the dropout rate is set to the same rate when using gradient dropout.

\subsubsection{Baseline Masking Methods}
Below we summarize the baselines considered in the our experiments.  
\newline
\textbf{GradFreeze+BottomTop}~\cite{Howard2018ulmfit}: Gradually unfreezes gradients during training from the bottom layer to the top layer after each epoch. 
\newline
\textbf{GradFreeze+TopBottom}~\cite{Howard2018ulmfit}: Gradually unfreezes gradients during training from the top layer to the bottom layer after each epoch.
\newline
\textbf{SFT}: Fine-tunes the whole network on the downstream task. In the proceeding results, this explicitly refers to XLM-R$_\text{Large}$.
\newline
\textbf{Unicoder}~\cite{huang2019unicoder}: Model trained with cross-lingual alignment using translation data. 
\newline
\textbf{FILTER}~\cite{fang2020filter}: As an upper bound on the expected performance, we include FILTER which too uses but cross-lingual alignment, but is a larger model than Unicoder as it uses XLM-R$_{\text{Large}}$. FILTER, is currently state of the art (SoTA) on the XGLUE benchmark.

% \textbf{LayerDrop}~\cite{fan2019reducing}: Randomly drops out self-attention blocks during training. 

\subsubsection{Our Gradient Masking Methods} Our proposed methods: \\
\textbf{GradDrop}: Randomly drops out gradients ($p=0.2$) on all layers for each batch.
\newline
\textbf{Layerwise GradDrop}: Randomly drops gradients ($p=0.2$) of a subset of layers for each mini-batch. 
\newline
\textbf{Anneal GradDrop}: Randomly drops gradients of weights (Anneal-GradDrop) or a subset of layers (Anneal-Layer-GradDrop) for each mini-batch, starting at a high gradient dropout rate ($p=0.9$) and finishing low ($p\approx 0$).
\newline
\textbf{GradDrop-Epoch}: Gradually unfreezes gradients randomly without replacement at each epoch until the whole network is unfrozen by the last epoch.
%Resevoir sampling~\cite{vitter1985random} can be used to uniformly sample without replacement. 
Further comparisons between the GDVs are in the supplementary material.

\section{Results}

In this section, we report the results of our proposed methods on the XGLUE benchmark tasks. We begin by discussing the zero-shot transfer results on sentence classification tasks.

\subsection{Sentence Classification Results}
\paragraph{Cross-lingual Natural Language Inference (XNLI)}
Table~\ref{tbl:xnli_detailed_results} shows the previous SoTA results on XNLI, our fine-tuned XLM-R$_{\text{Large}}$, {\em GradFreeze} (i.e., gradual unfreezing), GradDrop and its variants. Standard GradDrop outperforms its other variants and all prior SoTA fine-tuning methods, including gradient freezing. Our proposed methodology reports a 0.72\% increase in zero-shot accuracy for GradDrop compared to SFT.  
% Specifically, annealing the gradient dropout rate from $p=0.9$ to $p=0$ throughout training improves performance. 

\begin{table}[t]
    \begin{center}
        % \scriptsize
        \resizebox{1.\linewidth}{!}{
        \begin{tabular}[b]{lccccc|l}
        \toprule
         & \textbf{de} & \textbf{en} & \textbf{es} & \textbf{fr} & \textbf{ru} & \textbf{Avg.} \\
        \midrule
        FILTER~\cite{fang2020filter} & - & - & - & -  & - & 83.5 \\
        Unicoder~\cite{huang2019unicoder} & - & - & - & - & - & {83.5} \\
        \midrule
        XLM-R$_{\text{Large}}$ & 83.82 & 92.71 & 83.01 & 78.00 & 78.53 & {83.21} \\
        + GradFreeze-TopDown & 84.35 & 92.76 & 83.26 & 78.90 & 79.01 & 83.65 \\ % (+0.44)
        + GradFreeze-BottomUp & 79.75 & 90.27 & 81.31 & 75.18 & 75.56 & 80.41\\ % (-2.79)
         %+ LayerDrop & 82.77 & 92.62 & 82.70 & 77.38 & 80.78 & 83.25 \\ % (+0.04)
        \midrule
        + GradDrop  & 84.64${\dagger}$ & 92.84 & 83.26 & 78.57 & 79.30 & 83.41 \\ %  (+0.20)
        + Anneal Graddrop  & 74.36 & 93.13${\dagger}$ & 78.49 & 81.49 & \textbf{81.87} & 81.87 \\
        +Anneal Layer-GradDrop & 82.77 & 92.62 & 82.70 & 77.38 & 80.78${\dagger}$ & 83.25 \\
        + Layer-GradDrop & \textbf{84.95} & \textbf{93.55} & 84.08${\dagger}$ & 79.25${\dagger}$ & 79.43 & {\textbf{84.24}}  \\  
        + GradDrop-Epoch  & 83.65 & 92.78 & \textbf{84.14} & 78.58 & 79.42 & 83.73$^{\dagger}$ \\ %  (+0.42)
        \bottomrule
        \multicolumn{7}{l}{Top results are marked in {\bf bold}, while second best results are indicated with ${\dagger}$.}
        \end{tabular}
        }
        \caption{\textbf{Fine-Tuning XLM-R$_\text{Large}$ Results on News Classification. } Test Accuracy on English and Zero-Shot Results for German, Spanish, Russian and French.
        \label{tab:ft_news_classification}}
    \end{center}
   \vspace{-0.6cm}
\end{table}

\paragraph{News Classification}
Table~\ref{tab:ft_news_classification} shows the results on news classification where a category for news article is predicted and evaluated in 5 languages and trained on English. We find that both GradDrop and GradDrop-Epoch outperform the SoTA results (i.e FILTER) without any cross-lingual alignment techniques. We find that all GDVs outperform SFT of XLM-R$_{\text{Large}}$. We also find that \textit{gradual unfreezing} outperforms SFT and best performance is obtained only after 3 epochs, which corresponds to only 6 of 24 layers being unfrozen. This suggests that the news classification task is closely aligned to the learned features in the pretrained XLM-R$_{\text{Large}}$. We also note that Layer-GradDrop \emph{outperforms} FILTER by 1.26 percentage points.

\begin{table}[t]
    \begin{center}
        % \scriptsize
        \resizebox{1.\linewidth}{!}{
        \begin{tabular}[b]{lccc|l}
        \toprule
         & \textbf{de} & \textbf{en} & \textbf{fr} & \textbf{Avg.} \\
        \midrule
        FILTER ~\cite{fang2020filter} & - & - & - & 73.4 \\
        Unicoder ~\cite{huang2019unicoder} & - & - & - & {68.9} \\
        XLM-R$_{\text{Large}}$ & 70.10 & 70.83 & 68.52 & {69.82} \\
       \midrule
        +GradFreeze-TopDown & 71.85 & 72.16 & 69.03 & 71.01 \\
        +GradFreeze-BottomUp & 65.52 & 65.02 & 63.90 & 64.81\\
        \midrule
        +GradDrop  & 72.14 & 72.53 & 70.49 & 71.72 \\
        +Anneal-GradDrop  & 72.02 & 72.19 & 70.17 & 71.46 \\
        +Anneal Layer-GradDrop & 72.79 & 73.07${\dagger}$ & 71.05 & 72.31 \\
        +Layer-GradDrop  & 72.89${\dagger}$ & 72.88 & 71.23${\dagger}$ & 72.33$^{\dagger}$ \\
        +GradDrop-Epoch & \textbf{73.45} & \textbf{73.78} & \textbf{71.84} & {$\textbf{72.98}$} \\
        
        \bottomrule
        \end{tabular}
        }
        \caption{\textbf{XLM-R$_\text{Large}$ Zero-Shot Results on Question Answer Matching.} German, English \& French Test Accuracy.
        \label{tab:ft_qam}}
        \vspace{-3em}
    \end{center}
\end{table}

\begin{table}[!bt]
    \begin{center}
        % \scriptsize
        \resizebox{1.\linewidth}{!}{
        \begin{tabular}[b]{lccc|l}
        \toprule
         & \textbf{de} & \textbf{en} & \textbf{fr} & \textbf{Avg.} \\
        % \midrule
        % Encoder & Decoder & \multicolumn{6}{c}{} \\
        \midrule
        FILTER ~\cite{fang2020filter}  & - & - & - & 71.4 \\
        Unicoder ~\cite{huang2019unicoder} & - & - & - & {68.4} \\
        XLM-R$_{\text{Large}}$ & 69.57 & 71.95${\dagger}$ & 71.65 & {71.05} \\
        \midrule
        +GradFreeze-TopDown & 69.03 & 71.85 & 72.16 & 71.01 \\
        +GradFreeze-BottomUp & 66.02 & 69.37 & 70.28 & 68.56 \\
        \midrule
        +GradDrop & 69.53 & 71.89 & 71.60 & 71.01 \\
        +Anneal-GradDrop & 69.01 & 71.55 & 71.57 & 70.71 \\
        +Anneal Layer-GradDrop & 70.04 & 71.59 & 71.84${\dagger}$ & 71.16$^{\dagger}$ \\
        +Layer-GradDrop & \textbf{70.30} & \textbf{71.98} & \textbf{71.94} & {\textbf{71.39}} \\
        +GradDrop-Epoch & 70.12${\dagger}$ & 70.33 & 71.10 & 70.52 \\
        \bottomrule
        \end{tabular}
        }
        \caption{\textbf{Fine-Tuning XLM-R$_\text{Large}$ Results on Query-Ad Matching.} German, English and French Test Accuracy.
        \label{tab:ft_qadsm}}
        \vspace{-2em}
    \end{center}
\end{table}

\begin{table}[!bp]
    \begin{center}
        % \scriptsize
        \resizebox{1.\linewidth}{!}{
        \begin{tabular}[b]{lcccc|l}
        \toprule
         & \textbf{de} & \textbf{en} & \textbf{es} & \textbf{fr} & \textbf{Avg.} \\
        \midrule
        FILTER~\cite{fang2020filter} & - & - & - & - & 93.8 \\
        Unicoder ~\cite{huang2019unicoder} & - & - & - & - & {90.1} \\
        XLM-R$_{\text{Large}}$ & 85.23 & 93.65 & 88.70 & 89.35 &{89.23} \\
        \midrule
        + GradFreeze-TopDown & 85.13 & 93.15 & 87.03 & 88.15 & 88.33 \\
        + GradFreeze-BottomUp & 83.59 & 92.10 & 85.67 & 87.94 & 87.31 \\
        \midrule

        + GradDrop & 88.95${\dagger}$ & 94.90${\dagger}$ & 90.71${\dagger}$ & 91.35 & 91.46$^{\dagger}$ \\
        + Anneal Graddrop  & 88.78 & 94.38 & 90.18 & 91.48 & 91.18 \\
        + Anneal Layer-Graddrop & 88.48 & 94.48 & 89.19 & \textbf{92.09} & 91.06 \\
        + Layer-GradDrop  & \textbf{88.97} & \textbf{95.72} & \textbf{91.05} & 91.98${\dagger}$ & {\textbf{92.23}}  \\
        + GradDrop-Epoch & 88.75 & 94.55 & 90.55 & 90.80 & 91.15  \\
        %+ Layer-GradDrop-Epoch & 1 & 2 & 3 & 4 & 5 \\
        \bottomrule
        \end{tabular}
        }
        \caption{\textbf{Fine-Tuning XLM-R$_\text{Large}$ Results on Cross-lingual Adversarial Paraphrase Identification. } Test F1 score on English and (Zero-Shot) German, Spanish \& French.
        \label{tab:ft_pawsx_classification}}
    \end{center}
   \vspace{-0.6cm}
\end{table}

\begin{table}[t]
    \begin{center}
        % \scriptsize
        \resizebox{1.\linewidth}{!}{
        \begin{tabular}[b]{lcccc|l}
        \toprule
         & \textbf{de} & \textbf{en} & \textbf{es} & \textbf{nl} & \textbf{Avg.} \\
        \midrule
        FILTER ~\cite{fang2020filter} & - & - & - & - & 82.6 \\
        Unicoder ~\cite{huang2019unicoder} & - & - & - & - & {79.70} \\
        XLM-R$_{\text{Large}}$~\cite{Conneau2020xlm-r} & 72.27 & 92.74 & 76.44 & 81.00 & {80.61} \\
        XLM-R$_{\text{Large}}$ & 74.67$^{\dagger}$ & 93.15$^{\dagger}$ & 78.74$^{\dagger}$ & 82.10 & {82.16}  \\        \midrule
        + GradFreeze-TopDown & 68.52 & 91.33 & \textbf{78.87} & 75.84 & 78.64 \\
        + GradFreeze-BottomUp & 65.05 & 89.42 & 75.11 & 75.84 & 76.35 \\
        % + LayerDrop & 70.52 & 91.33 & 77.87 & 80.84 & 80.14 \\
        \midrule
        + GradDrop  & 74.20 & 91.23 & 76.22 & 79.18 & 80.21 \\
        + Anneal Graddrop  & 74.36 & 93.13 & 78.49 & \textbf{81.49} & 81.87$^{\dagger}$ \\
        %+ Anneal Layer GradDrop 81.02 & 96.78 &  78.79 & 82.89 & 84.87 \\
        + Anneal Layer GradDrop & 74.58 & 92.62 & 77.98 & 80.75 & 81.48 \\
        + Layer-GradDrop & 74.36 &  93.13 & 78.49 & 81.41 & 81.85 \\
        + GradDrop-Epoch & \textbf{79.47} & \textbf{94.95} & 73.83 & 81.39$^{\dagger}$ & {\textbf{82.41}}  \\
        %78.79 & 82.89
        %+ Layer-GradDrop-Epoch & 1 & 2 & 3 & 4 & 5 \\
        \bottomrule
        \end{tabular}
        }
        \caption{\textbf{Fine-Tuning XLM-R$_\text{Large}$ Results on Named Entity Recognition. } Test F1 score on English and Zero-Shot Results in German, Spanish and Dutch.
        \label{tab:ft_ner_classification}}
    \end{center}
   \vspace{-0.6cm}
\end{table}

\begin{table*}[t]
    \begin{center}
        % \scriptsize
        \resizebox{1.\linewidth}{!}{
        \begin{tabular}[b]{lcccccccccccccccccc|l}
\toprule
 & \textbf{en} & \textbf{ar} & \textbf{bg} & \textbf{de} & \textbf{el} & \textbf{es} & \textbf{fr} & \textbf{hi} & \textbf{it} & \textbf{nl} & \textbf{pl} & \textbf{pt} & \textbf{ru} & \textbf{th} & \textbf{tr} & \textbf{ur} & \textbf{vi} & \textbf{zh} & \textbf{Avg.} \\
\midrule
FILTER ~\cite{fang2020filter} & - & - & - & - & - & - & - & - & - & - & - & - & - & - & - & - & - & - & 81.6 \\
Unicoder ~\cite{huang2019unicoder}  & - & - & - & - & - & - & - & - & - & - & - & - & - & - & - & - & - & - & {79.6} \\
XLM-R$_{\text{Large}}$ (Ours) & 96.37 & 69.66 & \textbf{89.77} & 91.92$^{\dagger}$ & 87.99$^{\dagger}$ & 89.49 & \textbf{90.70} & 71.86 & 93.06 & 88.91 & 84.83$^{\dagger}$ & 90.47 & 86.55 & 57.44 & 72.71 & 64.09 & 57.82 & 63.27$^{\dagger}$ & {80.38} \\
\midrule
+ GradFreeze-Tog-TopDown  & 96.59 & 66.58 & 87.64 & 90.53 & 87.49 & 80.17 & 77.48 & 70.15 & 88.12 & 88.00 & 83.43 & 87.09 & 85.19 & 55.63 & 72.47 & 65.34 & \textbf{59.06} & 56.88 & 77.66 \\

+ GradFreeze-TopDown  & 96.80 & 65.33 & 88.40 & 90.40 & \textbf{89.63} & 81.41 & 83.25 & 70.12 & 90.16 & 88.18 & 83.09 & 87.19 & 85.20 & 55.73 & 72.26 & 66.96 & 58.50 & 57.85 & 78.11 \\

+ GradFreeze-BottomUp & 96.48 & 65.38 & 86.12 & 89.81 & 85.91 & 80.00 & 76.18 & 69.77 & 87.27 & 87.98 & 84.02 & 86.16 & 84.32 & 55.39 & 72.08 & 65.59 & 58.26 & 56.17 & 77.32 \\
\midrule

+ GradDrop & 96.57 & 72.54$^{\dagger}$ & 88.52 & \textbf{91.98} & 87.07 & \textbf{89.66} & 90.00 & 73.63 & 93.30$^{\dagger}$ & 88.80 & 84.60 & \textbf{90.86} & 87.27 & 58.80 & 74.74 & \textbf{70.94} & 57.65 & 60.95 & 81.00$^{\dagger}$\\

+ Anneal Graddrop  & 95.04 & 70.19 & 86.04 & 91.38 & 86.78 & 88.44 & 90.09 & 73.87 & 93.02 & 88.28 & 84.17 & 90.77 & 86.59 & 58.91$^{\dagger}$ & 74.01 & 69.37 & 57.40 & 60.13 & 80.72\\

% dev layer-graddrop 96.95 & 72.44 & 88.31 & 92.35 & 85.17 & 89.71 & 91.21 & 74.86 &  93.53 & 90.67 & 84.70  & 91.35 & 87.31 & 59.69 & 74.57 &  70.35 &  57.05 &  62.50 & 81.26
+ Anneal Layer-GradDrop & 96.41 & 72.19 & 87.91 & 91.40 & 86.59 & 89.33 & 89.85 & 73.28 & 92.94 & 88.67 & 84.29 & 90.72 & 87.03 & 58.62 & 74.59 & 70.18 & 57.17 & 60.52 & 80.65 \\

+ Layer-GradDrop & 97.01$^{\dagger}$ & \textbf{74.12} & 88.65$^{\dagger}$ & 91.67 & 86.38 & 89.56$^{\dagger}$ & 90.10$^{\dagger}$ & \textbf{74.82} & \textbf{93.38} & \textbf{89.03} &\textbf{84.94} & 90.81$^{\dagger}$ & \textbf{87.33} & \textbf{59.70} & \textbf{75.15} & 70.89$^{\dagger}$ & 58.10$^{\dagger}$ & \textbf{64.29} & {\textbf{81.42}} \\

+ GradDrop-Epoch & \textbf{97.03} & 72.21 & 88.55 & 91.87 & 86.57 & 88.73 & 89.57 & 74.43$^{\dagger}$ & 92.84 & 89.00$^{\dagger}$ & 84.36 & 90.73 & 86.69 & 58.34 & 75.46$^{\dagger}$ & 69.32 & 57.58 & 54.91 & 80.45 \\

% + Anneal Layer-GradDrop & \textbf{96.01} & \textbf{73.79} & \textbf{90.04} & \textbf{92.90} & \textbf{88.79} & \textbf{89.92} & \textbf{89.98} & \textbf{69.33} & \textbf{93.70} & \textbf{89.02} & \textbf{85.55} & \textbf{89.72} & \textbf{89.25} & \textbf{62.01} & \textbf{76.91} & \textbf{66.01} & \textbf{58.84} & \textbf{57.73} & \textbf{81.30}$^{\dagger}$  \\

%+ Layer-GradDrop-Epoch  & - & - & - & - & - & - & - & - & - & - & - & - & - & - & - & - & - & \\
\bottomrule
\multicolumn{20}{l}{The best performance obtained are marked in {\bf bold}, while the second best results are indicated with ${\dagger}$.}
\end{tabular}
}
\caption{\textbf{Fine-Tuning XLM-R$_\text{Large}$ Results on Part of Speech Tagging. } Test F1 score on English and Zero-Shot (17) languages.
\label{tab:ft_pos_classification}}
    \end{center}
   \vspace{-0.6cm}
\end{table*}

\paragraph{Question Answering Matching}
Table~\ref{tab:ft_qam} shows the zero-shot test accuracy on English, French and German for the Question-Answer Matching (QAM) results. The is involves predicting whether an answer answers a question correctly or not given a <question, answer> pair. We find that the GradDrop-Epoch variant outperforms other variants and improves significantly over SFT by 3.16\% and is only 0.42\% below FILTER. 
% We find that Anneal Layer-GradDrop performs best where the dropout rate begins at $0.9$ and finished at $0$. 

\subsection{Pairwise Classification }
 \paragraph{Query-Ad Matching Results}
In Query-Ad Matching (QADSM) task, we predict whether a advertisement is relevant to a query given an <query, advertisement> text input pair. We test performance on English and zero-shot test accuracy on French and German.
From Table~\ref{tab:ft_qadsm} we find that Layer-GradDrop outperforms the remaining GDVs, and is only 0.24\% accuracy percentage points below FILTER. 
 
 %which extends the Wikipedia portion of the PAWS~\cite{zhang2019paws} evaluation to more languages. We select 4 languages, including English, Spanish, French and German, from the original dataset and use them in XGLUE. Accuracy (ACC) of the binary classification is used as the metric. 

%\paragraph{Question-Answer Matching Results}

\paragraph{Cross-lingual Adversarial Paraphrase Identification}
The PAWS-X paraphrase identification dataset~\cite{yang2019paws} consists of English, Spanish, French and German languages for evaluation. From Table~\ref{tab:ft_pawsx_classification} we find that Layer-GradDrop is the best performing with 3\% improvements, and is competitive with FILTER ( $1.57$ F1 point difference).

\subsection{Structured Prediction Tasks}
\paragraph{Named Entity Recognition}
The Named Entity Recognition (NER) cross-lingual dataset is made up of CoNLL-2002 NER and
CoNLL-2003 NER~\cite{sang2003introduction}, covering  English, Dutch, German and Spanish with 4 named entities. From Table~\ref{tab:ft_ner_classification} we find that GradDrop-Epoch outperforms SFT, gradual unfreezing, the others, and is only 0.19\% points from FILTER. GradDrop outperforms SFT and is competitive with SoTA without additional parameters or training data. 

\paragraph{Part of Speech Tagging}
The Part of Speech (PoS) tagging dataset consists of a subset of the Universal Dependencies treebank~\cite{nivre2020universal} and covers 18 languages. From Table~\ref{tab:ft_pos_classification}, we find that all our GradDrop variants outperform SFT XLM-R$_{\text{Large}}$ and Layer-GradDrop is the best performing variant. Additionally, it is only 0.3\% average test accuracy points away from FILTER, the method that uses additional cross-lingual alignment training and pseudo-label knowledge transfer. Again, GradDrop does not rely on language alignment and only uses English language training data. We find that the largest improvements are made on Arabic, Urdu and Turkish (which shares approximately 30\% of its vocabulary with Arabic words written in Arabic).

\subsection{Sentence and Span Retrieval Tasks}
\paragraph{Web-Page Ranking} aims to predict whether a web page is relevant (1-5 ratings, ``bad'' to ``perfect'') to an input query and it is evaluated for 7 languages using the Normalized Discounted Cumulative Gain (nDCG). From Table~\ref{tab:ft_wpr}, we see that GradDrop-Epoch is the best performing gradient dropout variant, with Layer-GradDrop being 0.1 nDCG points below Layer-GraDrop and SFT being 1.29 points below GradDrop-Epoch. Moreover, GradDrop-Epoch is only 0.09 points from FILTER. 

 % Each labeled instance is a 4-tuple: <query, web page title, web page snippet, label>. The relevance label contains 5 ratings: Perfect (4), Excellent (3), Good (2), Fair (1) and Bad (0). We construct this dataset based on a commercial search engine. Normalize Discounted Cumulative Gain (nDCG) is used as the metric.

\begin{table}[t]
    \begin{center}
        % \scriptsize
        \resizebox{1.\linewidth}{!}{
        \begin{tabular}[b]{lccccccc|l}
        \toprule
         & \textbf{de} & \textbf{en} & \textbf{es} & \textbf{fr} & \textbf{it} & \textbf{pt} & \textbf{zh} & \textbf{Avg.} \\
        % \midrule
        % Encoder & Decoder & \multicolumn{6}{c}{} \\
        \midrule
        FILTER~\cite{fang2020filter} & - & - & - & - & - & - & - & 74.7 \\
        Unicoder ~\cite{huang2019unicoder} & - & - & - & - & - & - & - & {73.9} \\
        XLM-R$_{\text{Large}}$  & 76.91 & 77.78 & 75.67 & 74.60 & 68.18 & 77.53 & 62.58 & 73.32 \\
        \midrule
        +GradFreeze-TopDown  & 76.75 & 76.97 & 74.79 & 73.81 & 66.55 & 77.08 & 62.31 & 72.61 \\
        +GradFreeze-BottomUp & 73.42 & 73.58 & 74.01 & 72.84 & 67.04 & 75.13 & 62.18 & 71.17 \\
        \midrule
        +GradDrop  & 77.43 & 77.74 & 75.76 & 74.52 & 68.52 & 77.77 & 62.60 & 73.44 \\
        +Anneal-GradDrop  & 77.02 & 77.56 & 75.15 & 74.83 & 68.91 & 76.98 & 62.44 & 73.27 \\
        +Anneal Layer-GradDrop & 78.00 & 78.41 & 76.32 & 75.36$^{\dagger}$ & 69.29 & 78.74$^{\dagger}$ & 63.67 & 74.25 \\
        +Layer-GradDrop & 78.48$^{\dagger}$ & 78.83$^{\dagger}$ & 76.40$^{\dagger}$ & 75.12 & \textbf{70.00} & 78.65 & \textbf{64.08} & 74.51$^{\dagger}$ \\
        +GradDrop-Epoch & \textbf{78.93} & \textbf{78.85} & \textbf{76.70} & \textbf{75.61} & 69.33$^{\dagger}$ & \textbf{79.01} & 63.86 & {\textbf{74.61}} \\
        \bottomrule
        \multicolumn{9}{l}{\small Top results are in {\bf bold}, while second best results are indicated with ${\dagger}$.}
        \end{tabular}
        }
        \caption{\textbf{Fine-Tuning XLM-R$_\text{Large}$ Results on Web Page Ranking.} Normalized DCG on German, English, Spanish, French, Italian, Portuguese and Chinese.
        \label{tab:ft_wpr}}
    \end{center}
\end{table}

\begin{table}[t]
    \begin{center}
        % \scriptsize
        \resizebox{1.\linewidth}{!}{
        \begin{tabular}[b]{lccccccc|l}
        \toprule
        & \textbf{ar} & \textbf{de} & \textbf{en} & \textbf{es} & \textbf{hi} & \textbf{vi} & \textbf{zh} & \textbf{Avg.} \\ 
        \midrule
        XLM~\cite{lewis2019mlqa} & 54.8 & 62.2 & 74.9 & 68.0 & 48.8 & 61.4 & 61.1 & 61.6 \\
        FILTER~\cite{fang2020filter} & - & - & - & - & - & - & - & 74.7 \\
        Unicoder ~\cite{huang2019unicoder} & - & - & - & - & - & - & - & 66.0 \\
        XLM-R$_{\text{Large}}$ & 64.11 & 72.17 & 85.13 & 70.83 & 60.73 & 71.52 & 71.81 & {70.9} \\
        \midrule
        +GradFreeze-TopDown & 63.82 & 71.98 & 84.41 & 71.05 & 61.02 & 70.17 & 71.44 & 70.55 \\
        +GradFreeze-BottomUp & 61.29 & 70.48 & 84.02 & 69.98 & 60.79 & 69.88 & 71.05 & 69.64 \\
        \midrule
        +GradDrop & 64.91 & 72.66 & 85.47 & 71.00 & 60.98 & 71.90 & 72.12 & 71.29 \\
        +Anneal-GradDrop & 64.74 & 72.53 & 85.29 & 70.89 & 61.05 & 71.71 & 72.22 & 71.20 \\
        +Anneal Layer-GradDrop & 65.01 & 72.66 & 85.47 & 71.03 & 61.22 & 71.85 & 72.56 & 71.40 \\
        +Layer-GradDrop  & \textbf{66.09} & \textbf{73.60} & \textbf{87.02} & \textbf{72.17} & \textbf{61.65} & \textbf{72.50} & 72.59$^{\dagger}$ & \textbf{{72.25}} \\
        +GradDrop-Epoch & 65.60$^{\dagger}$ & 73.19$^{\dagger}$ & 86.32$^{\dagger}$ & 71.87$^{\dagger}$ & 61.58$^{\dagger}$ & 72.29$^{\dagger}$ & \textbf{72.75} & 72.01${^\dagger}$ \\
        \bottomrule
        \multicolumn{9}{l}{\small Top results are in {\bf bold}, while second best results are indicated with ${\dagger}$.}
        \end{tabular}%
        }
        \caption{\textbf{Cross-Lingual Transfer Results on MLQA.} F1 on Arabic, German, English, Spanish, Hindi, Vietnamese and \emph{Simplified} Chinese.
        \label{tab:ft_mlqa}}
    \end{center}
\end{table}

\begin{figure}[!bp]
\begin{subfigure}{.24\textwidth}
  \centering
  \includegraphics[width=1.\linewidth]{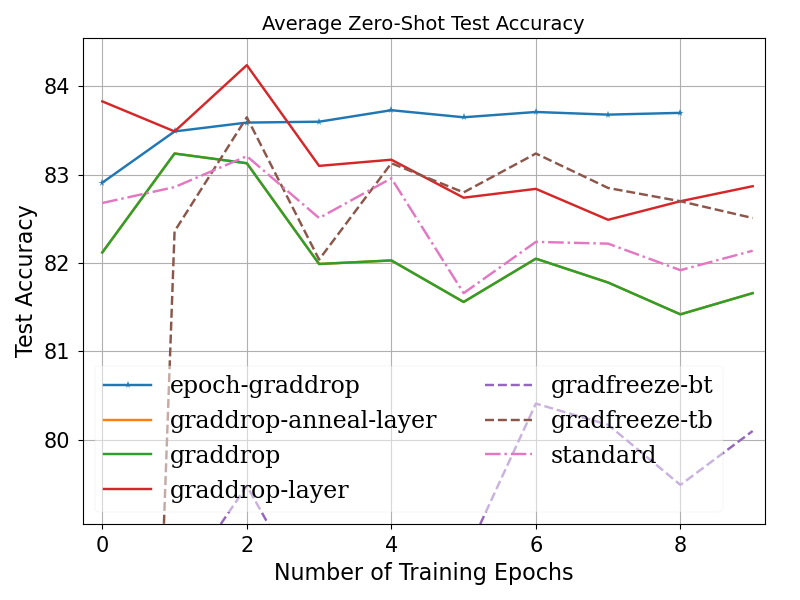}
  \caption{NC}
  \label{fig:nc}
\end{subfigure}%
\begin{subfigure}{.245\textwidth}
  \centering
  \includegraphics[width=1.\linewidth]{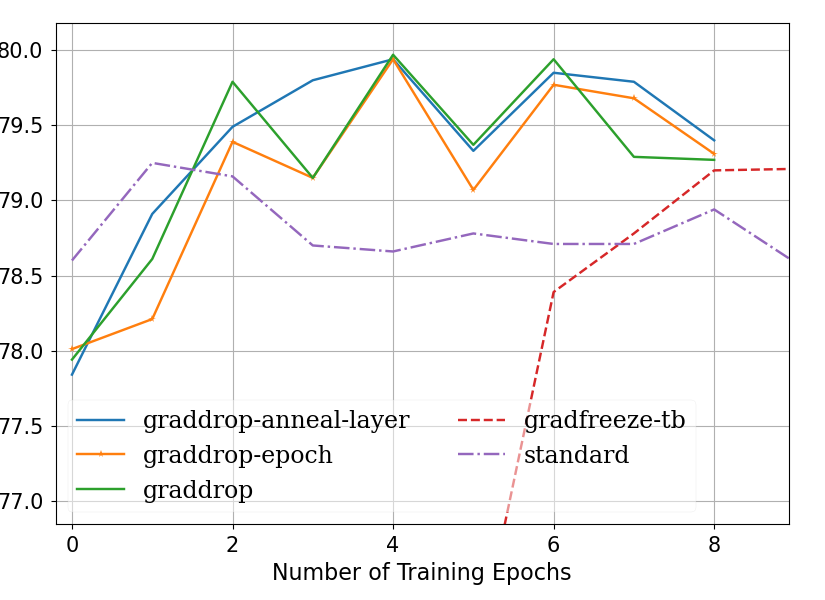}
  \caption{XNLI}
  \label{fig:xnli}
\end{subfigure}
\hfill
\begin{subfigure}{.24\textwidth}
  \centering
  \includegraphics[width=1.\linewidth]{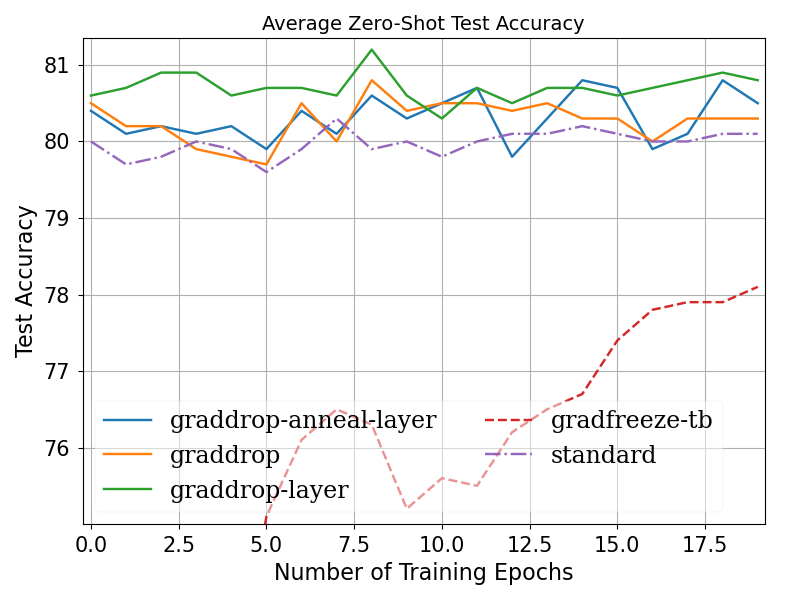}
  \caption{POS}
  \label{fig:pos}
\end{subfigure}%
\begin{subfigure}{.245\textwidth}
  \centering
  \includegraphics[width=1.\linewidth]{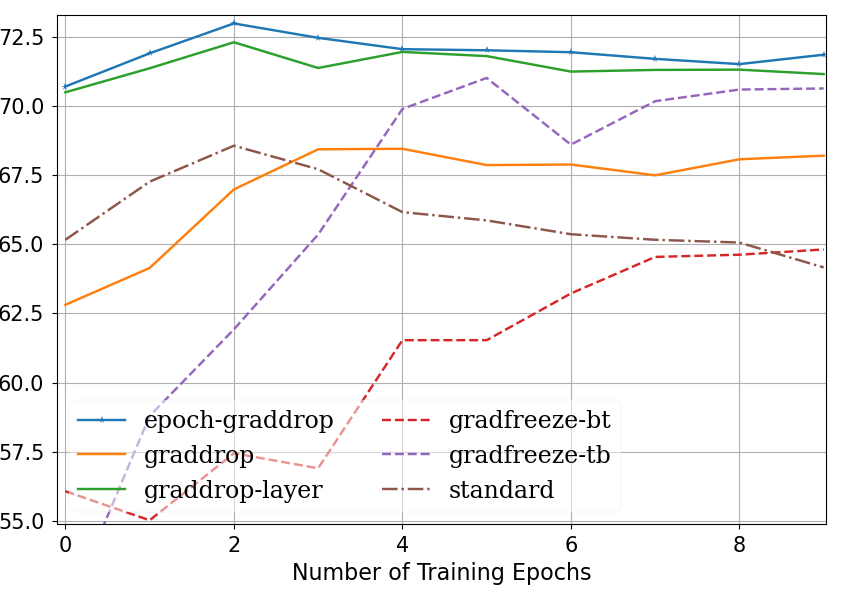}
  \caption{QAM}
  \label{fig:qam}
\end{subfigure}
\vspace*{-2mm}
\caption{\textbf{Test Performance Per Training Epoch.}}
\label{fig:stability}
\end{figure}

\begin{table*}[ht]
\begin{center}
\vspace*{-4mm}
    % \scriptsize
    \resizebox{1.\linewidth}{!}{
    \begin{tabular}[b]{l|cc|cccccccccl}
    \toprule
    \textbf{Models} & \textbf{Translation} & \textbf{\#Params} & \textbf{XNLI} & \textbf{NC} & \textbf{NER} & \textbf{PAWSX} & \textbf{POS} & \textbf{QAM} & \textbf{QADSM} & \textbf{WPR} & \textbf{MLQA} & \textbf{Avg.} \\ 
    \midrule
    
%NER	  POS   NC    MLQA  XNLI  PAWS-X   QADSM  WPR   QAM   AV
%79.7  79.6  83.5  66.0  75.3  90.1	   68.4	  73.9  68.9  76.1
    
    %FILTER~\cite{fang2020filter} & Yes & 550M & 	  83.6 & 83.5 & 80.49 & 89.23 & 81.3 & 81.6 & 73.4 & 71.4 & 74.7 & 79.91 \\	
    M-BERT~\cite{liang2020xglue} & Yes & 550M & 	66.3 & 82.7 & 78.2 & 87.2 &  74.7 & 66.1 & 64.2 & 73.5 & 60.7 & 72.6 \\	
    FILTER+Self-Teaching~\cite{fang2020filter} & Yes & 550M & 83.9 & 83.5 & 82.6 & 93.8 & 81.6 & 73.4 & 71.4 & 74.7 & 76.2 & 80.1 \\
    XLM-R$_{\text{Large}}$-T~\cite{fang2020filter} & Yes & 550M & 82.6 & - & - & - & - & - & - & - & - & - \\
    \midrule
    Unicoder~\cite{huang2019unicoder} & No & 255M & 75.3 & 83.5 & 79.70 & 90.1 & 79.6 & 68.9 & 68.4 & 73.9 & 66.0 & 76.1 \\
    XLM-R$_{\text{Large}}$~\cite{conneau2019unsupervised} & No & 550M & 80.1 & - & - & - & - & - & - & - & - & \\
    XLM-R$_{\text{Large}}$~\cite{fang2020filter} & No & 550M & 79.2 & 83.2 & - & - & - & - & - & - & - & - \\
  
    XLM-R$_{\text{Large}}$ (Ours) & No & 550M & 79.25 & 83.21 & 80.61 & 89.23 & 80.38 & 69.82 & 71.05 & 73.27 & 70.21 & 77.45  \\  
    \midrule
    \midrule
    +GradFreeze-TopDown & No & 550M & 79.23 & 83.65 & 78.64 & 88.33 & 78.11 & 71.01 & 71.01 & 72.61 & 70.55 & 77.02 \\
    +GradFreeze-BottomUp & No & 550M & 78.07 & 80.41 & 76.35 & 87.31 & 73.32 & 64.81 & 68.56 & 71.17 & 69.64 & 74.40 \\
    \midrule
    +GradDrop & No & 550M & \textbf{79.97} & 83.41 & 80.21 & 91.46$^{\dagger}$ & 81.00$^{\dagger}$ & 71.72 & 71.02 & 73.44 & 71.29 &  78.17 \\
    +Anneal-GradDrop & No & 550M & 79.58 & 81.87 & 81.87 & 91.18 & 80.72 & 71.46 & 70.71 & 73.27 & 71.20 & 77.98 \\

    +Anneal-Layer-GradDrop & No & 550M & 79.94$^{\dagger}$ & \textbf{84.24} & 81.48 & 91.06 & 80.88 & 72.31 & 71.16$^{\dagger}$ & 74.25 & 71.40 & 78.52 \\
    +Layer-GradDrop & No & 550M & 79.22 & 83.73$^{\dagger}$ & 81.85$^{\dagger}$ & \textbf{92.23} & \textbf{81.42} & \textbf{72.33} & \textbf{71.39} & 74.51$^{\dagger}$ & \textbf{72.55} & \textbf{{78.77}}  \\
    +GradDrop-Epoch & No & 550M & 79.94$^{\dagger}$ & 83.73$^{\dagger}$ & \textbf{82.41} & 91.15 & 80.45 & 72.98$^{\dagger}$ & 70.52 & \textbf{74.61} & 72.01$^{\dagger}$ & {{78.64}}$^{\dagger}$ \\
    \bottomrule
    \multicolumn{13}{l}{The best performance obtained are marked in {\bf bold}, while the second best results are indicated with ${\dagger}$.}
    \end{tabular}
    }
    \vspace*{-4mm}
    \caption{\textbf{Zero Shot Cross-Lingual Performance Per Task and Overall Average Score (Avg.).}
    \label{tab:ft_all}}
\end{center}
\vspace*{-4mm}
\end{table*}

\paragraph{Multilingual Question Answering}
We use MLQA~\cite{lewis2019mlqa} for the a
multilingual machine reading comprehension task,
which contains QA annotations labeled in 7 languages, including English, Arabic, German, Spanish, Hindi, Vietnamese and Chinese. Again, we find that Layer-GradDrop and GradDrop-Epoch are the best performing GDVs. Layer-GradDrop increases F1 by 1.35 over SFT, while being 2.35 below FILTER. 

%F1 score of the predicted answers is used as the metric \textbf{FINISH THIS !}

\subsection{Convergence and Stability Analysis}
We also analyse the stability of different GDVs, compared to SFT and gradual unfreezing in Figure~\ref{fig:stability}. In Figure~\ref{fig:nc}, the best test performance is found after 3 epochs for all GD variants. On further inspection, fine-tuning with GradDrop-Epoch maintains test performance for further training epochs while SFT decreases as the model begins to overfit. This can be attributed to a reduction in the number of parameters being trained at any given epoch. In the remaining 3 tasks (XNLI, POS and QAM), GradDrop variants maintain a stable test performance over training epochs. 

\begin{figure}[ht]
  \centering
  \includegraphics[width=\linewidth]{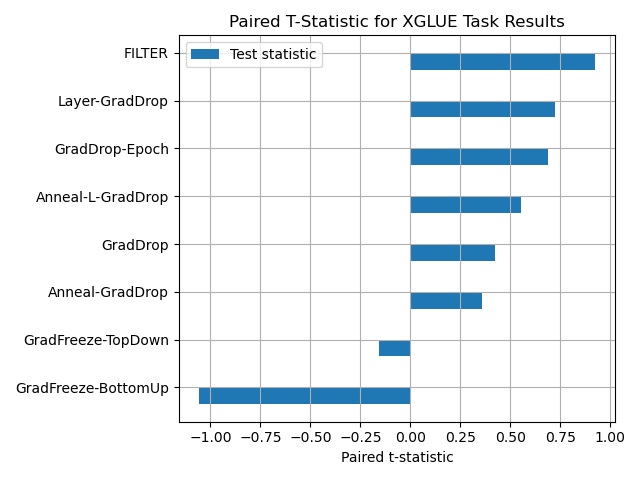}
\caption{\textbf{Test Statistic Across XGLUE Results}}
\label{fig:test_stat}
\end{figure}

\iffalse
\begin{figure}[ht]
\begin{subfigure}{.5\textwidth}
  \centering
  \includegraphics[width=.85\linewidth]{images/perf_difference/pos_large_labels.png}
  \label{fig:pos_lang}
\end{subfigure}
\hfill
\begin{subfigure}{.5\textwidth}
  \centering
  \includegraphics[width=.85\linewidth]{images/perf_difference/xnli_paper.png}
  \label{fig:xnli_lang}
\end{subfigure}
\caption{\textbf{Test Performance Increase By Language in POS (top) and XNLI (bottom).} Red bars indicate accuracy increases or decreases, while blue indicates the fractional increase of GradDrop over standard fine-tuning.}
\label{fig:perf_by_lang}
\end{figure}
\fi

\subsection{XGLUE Understanding Score}
Finally, we show the average task \textit{understanding} score for our GradDrop variants and previous baselines in Table~\ref{tab:ft_all}. We find that GradDrop-Epoch and Layer-GradDrop are two methods which consistently outperform the remaining GradDrop variants, SFT and in some cases, FILTER which uses translation data. To our knowledge, Layer-GradDrop sets a SoTA results on XGLUE for methods which \emph{do not} use \textit{translate-train} or translation language model \textit{cross-lingual} alignment pretraining. Additionally, Layer-GradDrop is only 1.4 understanding score points from FILTER with their self-teaching loss. Figure~\ref{fig:test_stat} shows the two-sided pairwise t-test between the zero-shot task performance of our fine-tuned XLM-R$_{\text{Large}}$ and each of our proposed GDVs. We see that again GradDrop-Epoch and Layer-GradDrop has the highest test statistic across 8 tasks.

\begin{figure}[h!]
  \centering
  \includegraphics[width=1\linewidth]{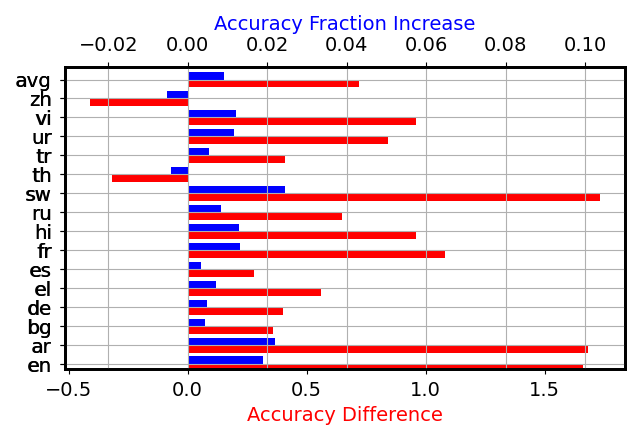}
  \label{fig:xnli_lang}
  %\vspace{-2em}
\caption{\textbf{Test Performance Increase By Language in XNLI.} Red bars indicate accuracy increases or decreases, while blue indicates the fractional increase of GradDrop over standard fine-tuning.}
\label{fig:perf_by_lang}
\end{figure}

\paragraph{Discussion}
% are two different task types i.e structured prediction and sentence pair classification
Lastly, we inspect what languages do GDVs improve performance the most when compared to SFT. We analyse XNLI which includes well-resourced and under-resourced languages in the evaluation set. Figure~\ref{fig:perf_by_lang} shows how our best performing GDVs increase over SFT and which languages we mostly attribute to the increase in average score.
%We find that there is significant performance improvements in Urdu, Turkish, Thai, Hindi and Arabic. In contrast, the performance of well-resourced languages such as English, Spanish, French, German and Italian are relatively minuscule. Likewise in XNLI, 
We find that biggest gains are made on Swahili and Arabic. We conclude that GradDrop improves performance on under-resourced languages in particular. We posit that this may be because GradDrop forces the model to be robust to static gradients during training on English only, reducing the effects of overfitting to the English language.
% which is destructive to the alignment learned from the translation LM in XLM-R$_{\text{Large}}$.

\section{Conclusion}
In this paper, we proposed GradDrop and its multiple variants, showing that these variants can outperform standard fine-tuning of cross-lingual pretrained transformers. Specifically, epochwise- and layerwise- gradient dropout consistently outperform standard fine-tuning, gradual unfreezing and other gradient dropout variants. Additionally, it is competitive against SoTA methods that use translation data, cross-lingual alignment pretraining and self-distillation. We also find that gradient dropout particularly improves fine-tuning performance for under-resourced languages.  

\bibliography{main}
\bibliographystyle{IEEEtran}

\end{document}